\documentclass{article}
\usepackage{libertine} %
\usepackage[letterpaper,margin=1in]{geometry}
\usepackage{style}

\usepackage{xcolor}
\usepackage[unicode=true]{hyperref}       %
\definecolor{mydarkblue}{rgb}{0,0.08,0.45}
\hypersetup{ %
    pdftitle={},
    pdfauthor={},
    pdfsubject={},
    pdfkeywords={},
    pdfborder=0 0 0,
    pdfpagemode=UseNone,
    colorlinks=true,
    linkcolor=mydarkblue,
    citecolor=mydarkblue,
    filecolor=mydarkblue,
    urlcolor=mydarkblue,
    pdfview=FitH
}

\usepackage{microtype}
\usepackage{graphicx}
\usepackage{subfigure}
\usepackage{booktabs} %
\usepackage{natbib}

\usepackage{hyperref}

\def\shownotes{1}  \ifnum\shownotes=1
\usepackage{todonotes}
\else
\usepackage[disable]{todonotes}
\fi

\usepackage{amsmath}
\DeclareMathOperator*{\argmax}{arg\,max}
\DeclareMathOperator*{\argmin}{arg\,min}

\usepackage{amsmath}
\usepackage{amssymb}
\usepackage{mathtools}
\usepackage{amsthm}

\usepackage[capitalize,noabbrev]{cleveref}

\title{Limitations of Neural Collapse\\for Understanding Generalization in Deep Learning}
\author{
Like Hui\\
\texttt{\sf lhui@ucsd.edu}
\and
Mikhail Belkin\\
\texttt{\sf mbelkin@ucsd.edu}
\and
Preetum Nakkiran\\
\texttt{\sf preetum@ucsd.edu}
\and
Hal{\i}c{\i}o{\u g}lu Data Science Institute\\
University of California San Diego
}
\date{}

\begin{document}

\maketitle

\begin{abstract}
The recent work of \citet*{Papyan24652} presented an intriguing ``Neural Collapse'' phenomenon,
showing a structural property of interpolating classifiers in the late stage of training.
This opened a rich area of exploration studying this phenomenon.
Our motivation is to study the upper limits of this research program:
How far will understanding Neural Collapse take us in understanding deep learning?
First, we investigate its role in generalization.
We refine the Neural Collapse conjecture into two separate conjectures:
collapse on the train set (an optimization property) and collapse on the test distribution (a generalization property).
We find that while Neural Collapse often occurs on the train set,
it does not occur on the test set.
We thus conclude that Neural Collapse is primarily an optimization phenomenon,
with as-yet-unclear connections to generalization.
Second, we investigate the role of Neural Collapse in representation learning.
We show simple, realistic experiments where more collapse leads
to \emph{worse} last-layer features, as measured by transfer-performance on a downstream task.
This suggests that Neural Collapse is not always desirable for
representation learning, as previously claimed.
Finally, we give preliminary evidence of a ``cascading collapse'' phenomenon, 
wherein some form of Neural Collapse occurs not only for the last layer, but in earlier layers as well.
We hope our work encourages the community to continue the rich line of Neural Collapse research,
while also considering its inherent limitations.
\end{abstract}

\section{Introduction}
\label{intro}
In science, and in deep learning, novel empirical observations often
catalyze deeper scientific understanding \citep{kuhn}.
When faced with a new or surprising experiment, we can then try to understand the
phenomenon more precisely: How universal is the behavior?
In what settings does it hold? Can we describe it quantitatively?
What does it teach us more generally?
This overall roadmap for understanding
---from observations to quantitative conjectures \& laws--- has a long history of success in
the natural sciences, and has also enjoyed recent successes in deep learning.
\begin{wrapfigure}{r}{0.4\textwidth}
\centering
\vspace{-0.2in}
\includegraphics[width=0.85\linewidth]{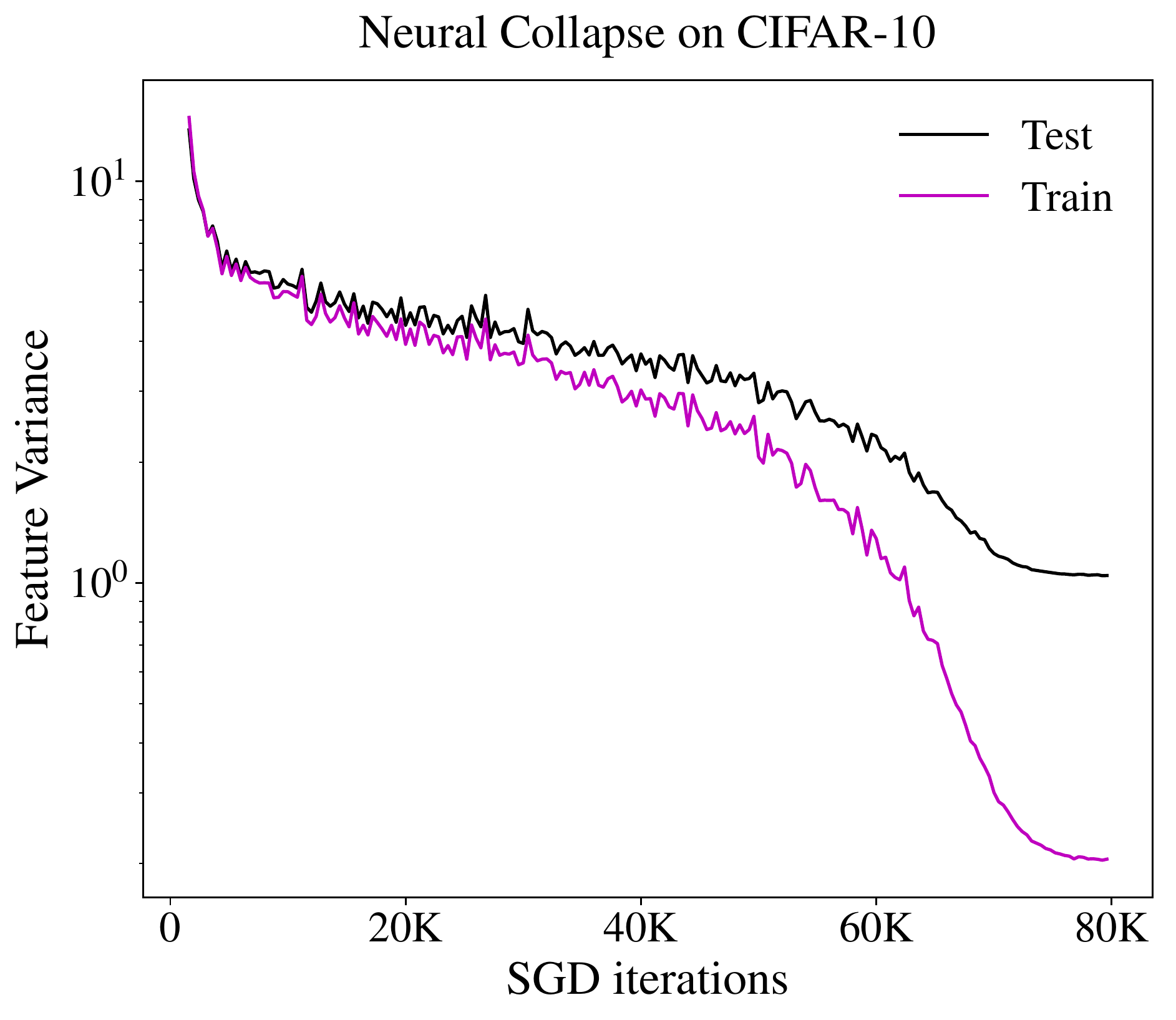}
\caption{{\bf Failure of Test Collapse.}
Neural Collapse for ResNet18 on CIFAR-10.
Collapse occurs on the train set, but not on test.}
\label{fig:train-test-cifar}
\vspace{-0.3in}
\end{wrapfigure}

The recent ``Neural Collapse'' work of
\citet*{Papyan24652} initiated another instance of such a research program in
understanding deep learning.
Their work presented a new experimental observation, along with a partial characterization.
At a high level, Neural Collapse conjectures
several structural properties of deep neural networks 
when trained past the point of $0$ classification error on the train set.
Their weakest conjecture--- and thus the one most likely to be true---
is ``variability collapse (NC1).''
Variability collapse proposes, informally, that when a deep network is trained on a $k$-way classification task,
the last-layer representations converge to $k$ discrete points.
This is apriori surprising, since this internal structure is in no way
required to achieve low train loss and high test performance: there exist networks
with identical decision boundaries which do not satisfy collapse.
However, our standard training methods (Stochastic Gradient Descent and variants)
on standard architectures and datasets empirically seem to satisfy some
form of collapse, as demonstrated in \citet{Papyan24652}.
This work has since inspired many
follow-up works investigating this phenomenon, both theoretically
and empirically.

A motivating factor in this research program is the
belief that Neural Collapse is not an isolated phenomenon,
but rather is deeply connected to
other important and unsolved aspects of deep learning---
in particular \emph{generalization}.
The problem of generalization, informally, is 
the study of why a model trained on a finite set of samples
has good performance on out-of-sample inputs.
Although this is not apriori related to Neural Collapse,
the original work proposes that collapse
``confers important benefits,
including better generalization performance, better robustness, and better interpretability.''
And it is stated as a hypothesis that
``the benefits of the
interpolatory regime of overparametrized networks are directly
related to Neural Collapse'' \citep{Papyan24652}.
This postulated connection between Neural Collapse and generalization
is implicit in many of the follow-up works as well,
and motivates studying collapse as a phenomenon.

However, the nature of the connection between Neural Collapse
and generalization remains muddled.
Some works argue they are closely related \citep{poggio2020explicit, banburski2021distribution},
while others cast some doubt \citep{elad2020another, zhu2021geometric}.
There are at least two reasons for this confusion in the literature:
First, it is often not clear whether Neural Collapse
refers to a phenomenon on the train set, or on the test set.
The behaviors most relevant to generalization occur on the test set,
and yet most experiments and theorems consider only the train set.
Second, the Neural Collapse conjectures do not precisely specify
the role of the sample size, and thus it is not always clear
how to connect to generalization--- where sample size
is fundamental.
This ambiguity is especially problematic because some natural
ways to extend the Neural Collapse conjecture
to the test set turn out to be impossible to satisfy,
as we will describe.

\paragraph{Our Contributions.}
We study the powers and limitations of the Neural Collapse (NC) research program
in understanding deep learning. That is, we study what NC can explain about deep learning,
and what it cannot.
To do so, we clarify several ambiguities surrounding the Neural Collapse conjectures,
and carefully investigate its relation to both generalization and optimization.
Specifically:
\begin{enumerate}
  \item We propose more precise versions of the Neural Collapse conjectures (``variability collapse''),
  stating different versions for the train set and the test set,
  with both ``strong'' and ``weak'' forms.
  \textbf{(\cref{sec:definitions})}
  \item We discuss the theoretical feasibility of these different conjectures. 
  As we will see, strong test collapse is extremely unlikely, while weak test collapse is in principle possible but does not occur in practice.
  \textbf{(\cref{sec:feasibility})}
  \item We empirically confirm the finding 
  of \citet{Papyan24652},
  that train-collapse occurs in many realistic settings.
  However, we find that test-collapse does not occur.
  \textbf{(\cref{sec:nc-experiments})}
  \item We show several settings where increasing train-collapse
  is anti-correlated with test performance,
  in both on-distribution and transfer-learning settings.
  This demonstrates that train-time neural collapse
  is not always desirable--- and indeed, can be counterproductive--- for many kinds of generalization.
  \textbf{(\cref{sec:feature-learning}}).
  \item  As an addendum, we present a preliminary proposal of how Neural Collapse
    could be extended to describe collapse of not only the final layer, but of earlier layers as well---a phenomena we term ``cascading collapse.''
    We leave fully investigating this as an interesting area for future work.
    \textbf{(\cref{sec:cascade})}.
\end{enumerate}

We thus conclude that Neural Collapse is primarily an \emph{optimization} phenomenon, and not a  \emph{generalization} one.

\subsection{Related Works}
The Neural Collapse phenomenon was originally presented in \citet{Papyan24652}, and led to a series of follow-up works investigating and extending it.
Many of the subsequent works develop simplified models
in which Neural Collapse on the train set
can be theoretically proven and understood.
For example, \citet{fang2021exploring} develops a ``layer-peeled'' model of training, and explores neural collapse in class-imbalanced settings.
\citet{mixon2020neural} proposes an alternate simplification,
an ``unconstrained features'' model, in which train collapse also occurs. \citet{wojtowytsch2020emergence} and \citet{zhu2021geometric} also investigate the train collapse under unconstrained features model. Several works \citet{poggio2020explicit, poggio2020implicit, rangamani2021dynamics, Han2021NeuralCU} examine the Neural collapse with the square loss under different settings. Specifically, \citet{poggio2020explicit, poggio2020implicit} give theory which predicts the properties of neural collapse for homogeneous, weight-normalized networks. \citet{rangamani2021dynamics} proves that quasi-interpolating solutions obtained by gradient descent in the presence of weight decay have Neural collapse properties. 
\citet{Han2021NeuralCU} proposes a generic decomposition of the MSE loss which, under certain assumptions,
results in a simplified dynamical description
(the ``central path'') which exhibits neural collapse on the train set. \citet{}
\citet{Lu2020NeuralCW} extend theoretical analysis of neural collapse to the cross-entropy loss (while previous works mainly considered MSE loss).
They prove neural collapse on the train set in the ``unconstrained features'' setting. \citet{ergen2021revealing} reformulate the last-layers of networks to convex formulations and give an explanation 
of Neural Collapse properties.
There are also other works \citep{ji2021gradient, ji2021unconstrained} investigating Neural Collapse under different settings.

However, all the above papers present results for Neural Collapse on the train set. \citet{Han2021NeuralCU} gives a preliminary experiment (see Figure 12 in \citet{Han2021NeuralCU}) on the test set collapse. One of the few papers focusing on collapse at test time is \citet{galanti2021role}.
Their work
considers feature variance for test samples as well.
However, the results of \citet{galanti2021role} do not satisfy our definition of test-collapse,
as we define it.
Specifically, to be considered a ``collapse'', we require
the collapse to occur for \emph{finite sample size $n$},
but in the limit of SGD steps $t \to \infty$.
We elaborate on this important point in Section~\ref{sec:definitions}.
In particular, \citet{galanti2021role} do not demonstrate test collapse in the
relevant regime according to our definitions.

\subsection{Notation}
Let $\mathcal{X}$ be the input space, and $\mathcal{Y}$ be the label space.
We consider multi-class classification problems,
where $\mathcal{X} = [k]$ for some $k \in \N$.
Let $\cD$ be the target distribution over $\mathcal{X} \x \mathcal{Y}$.
\emph{Training procedures}\footnote{We can consider randomized training procedures by allowing an additional random string as input.
We omit this randomness throughout, for notational clarity.}
are functions 
which map a train set $S \in (\mathcal{X} \x \mathcal{Y})^n$ and an iteration count $t \in \N$
and to a model $f$.
In this work, we will always consider
Stochastic-Gradient-Descent (SGD)-based training procedures,
where $t$ is the number of SGD steps.
For a fixed train set $S$ of size $n$, let
$f_{S}^t$ denote the model output by the training procedure after $t$ iterations.
So $\textsf{Train}: (S, t) \mapsto f_{S}^t$,
where $\textsf{Train}$ denotes the training procedure.
For a given model $f_{S}^t$,
let the \emph{last-hidden-layer feature map} be denoted
$$
h_{S}^t: \mathcal{X} \to \R^d.
$$
This is the feature-map induced by the trained model,
as a map from inputs into $\R^d$.

\section{Defining Neural Collapse}
\label{sec:definitions}
We first define two kinds of Neural Collapse:
on the train set, and on the test set.
Our definitions naturally extend the definitions in \citet{Papyan24652}, 
but are more precise since we
explicitly include the train/test distinction, and the dependency on training iterations $t$ and train samples $n$.
This is essential to describe the relevant asymptotic limits in the ``collapse''.

Throughout this work, we focus only on the first, and weakest, 
conjecture from \citet{Papyan24652}: ``NC1 (Variability Collapse).''
The subsequent conjectures (NC2-4) are particularly meaningful only if NC1 is true.
When we refer to ``neural collapse'' in this work, we specifically are referring to ``variability collapse.''
We first define collapse on the train set, which follows closely
the definition in \citet{Papyan24652}.

\begin{definition}[Train-Collapse]
\label{def:nc-train}
For a particular train set $S$,
we say a training procedure $T$
exhibits
\emph{Train-Collapse on $S$} if
there exists some distinct $\mu_1, \mu_2, \dots, \mu_k \in \R^d$ such that
$$
\forall (x_i, y_i) \in S: \quad \lim_{t \to \infty} h_{S}^t(x_i) = \mu_{y_i}
$$
\end{definition}

That is, the trained network converges to representations
such that all train points of class $k$ get embedded 
to a single point $\mu_{k}$ (called the ``class means'' in \citet{Papyan24652}).
The conjecture below then states conditions under which Train-Collapse occurs.
This conjecture is meant to capture the original NC1 conjecture of \cite{Papyan24652}, which was demonstrated empirically across many settings.

\begin{conjecture}[Train-Collapse Conjecture, informal]
\label{conj:nc-train}
For all train sets $S$ containing at least two distinct labels,
and all training procedures $T$ corresponding to SGD on ``natural''
sufficiently-deep and sufficiently-large neural network architectures:
\emph{$T$ exhibits Train-Collapse on $S$}.
\end{conjecture}

Crucially, we state Conjecture~\ref{conj:nc-train} for train sets of \emph{all sizes}.
This dependency on train set size is implicit, but omitted from \citet{Papyan24652}---
it will become especially important when we discuss test-collapse.
This behavior is called a ``collapse'' because regardless of the train set size,
any big-enough network will converge to this discrete limiting structure.
We replicated this finding in most of our experiments.
However, for completeness we acknowledge that
this conjecture does not hold fully universally,
and there are subtleties in practice\footnote{
For example, we found in some settings training variability does not collapse to negligible value, such as CIFAR-10 and STL-10 dataset with VGG architectures (see Figure \ref{fig:train_vs_test}) .
In some preliminary experiments we also found  that adding stochasticity (such as dropout noise) often accelerated collapse,
which is consistent with the theoretical model in \cite{Papyan24652}.}.
Nevertheless, we believe the NC1 conjecture captures the right qualitative behavior in many realistic settings.

We also acknowledge that Conjecture~\ref{conj:nc-train},
while more precise than the conjectures in \citet{Papyan24652},
is still not fully formal.
For example, it only applies to ``natural'' architectures and not all architectures,
and does not quantify what ``sufficiently large'' means.
This restriction to ``natural'' architectures is a known obstacle to formalism in deep learning theory (e.g. \citet{preetum-thesis})
and is necessary to avoid pathologies such as \citet{abbe}.
Nevertheless, our definitions take a step towards greater formalism, and this precision will be useful in understanding connections
to generalization. Refining our definitions and conjectures further is an area for future work.

The notion of train-collapse described above
(and in \citet{Papyan24652}) is an \emph{optimization} notion:
it involves only behavior of a model on its train set, and not behavior at test time.
Thus, it is a priori unclear whether this notion is related to generalization aspects of models.
To explore this, we first extend the definition of Neural Collapse to the test set, and then investigate whether
this test-collapse occurs in practice.
There are two natural ways to extend the notion of 
collapse to the test distribution: a ``weak'' way and a ``strong'' way.
Weak-collapse requires only that test points embed as
\emph{one of} $k$ discrete points $\mu_1, \mu_2, \dots \mu_k$,
without requiring that all points of class $i$ map to $\mu_i$.

\begin{definition}[Weak Test-Collapse]
\label{def:nc-test}
A training procedure $T$ 
exhibits
\emph{Weak Test-Collapse}
on distribution $\cD$ if
for all sample sizes $n \in \N$, the following holds with probability $1$
over sampling $S \sim \cD^n$:
there exists some distinct $\mu_1, \mu_2, \dots, \mu_k \in \R^d$ such that
$$
\text{with prob $1$ over $(x, y) \sim \cD$}: \quad
\lim_{t \to \infty} h_{S}^t(x) \in \{\mu_i\}_{i \in [k]}
$$
\end{definition}

Strong-collapse, on the other hand, requires that test points $x$ map to their ``correct'' embedding point $\mu_i$, where $i$ is the Bayes-optimal class for $x$.

\begin{definition}[Strong Test-Collapse]
\label{def:nc-test-strong}
A training procedure $T$ 
exhibits
\emph{Strong Test-Collapse} 
on distribution $\cD$ if
for all sample sizes $n \in \N$, the following holds with probability $1$
over sampling $S \sim \cD^n$:
there exists some distinct $\mu_1, \mu_2, \dots, \mu_k \in \R^d$ such that
$$
\text{with prob $1$ over $(x, y) \sim \cD$}: \quad
\lim_{t \to \infty} h_{S}^t(x) = \mu_{y^*(x)}
$$
where $y^*(x) := \argmax_y p_\cD(y | x)$ is the Bayes-optimal classification under distribution $\cD$.
\end{definition}

There are several important differences
between the notions of test-collapse and train-collapse.
First, for test-collapse we require that the train set $S$ is not arbitrary, but sampled from some distribution $\cD$.
And we check for limiting behavior with respect to 
\emph{new} samples from $\cD$, as opposed to train samples from $S$.
However, both train and test collapse require the collapse to occur
\emph{for all finite sample sizes $n$}, letting only time $t \to \infty$.
This is the meaningful asymptotic,
since taking limit of samples $n \to \infty$
would obscure almost all aspects of learning,
which is most interesting at finite-sample sizes.

\subsection{Remarks on Feasibility}
\label{sec:feasibility}
With the above definitions,
we can see that
strong test-collapse is too strong a property to apply in realistic settings.
We discuss this infeasibility here,
and then corroborate this with experiments in the following section.

\paragraph{Infeasibility of Strong Test-Collapse.}
First, note that both train-collapse and test-collapse
definitions require that collapse occurs for all train set sizes $n \in \N$.
This property is easy to satisfy for train-collapse, but is an extremely strong property for test-collapse.
In particular, the ``strong'' form of test collapse (Definition~\ref{def:nc-test-strong}) is too strong to hold
in practice:
it implies that
a Bayes-optimal classifier can be extracted from the trained model features, even if the model is trained on only e.g. $n=10$ samples.
This is because, according to Definition~\ref{def:nc-test-strong},
the representation must map test inputs to their ``correct'' cluster,
and thus the correct label can be extracted from the cluster identity.

However, the ``weak'' form of NC1-test (Definition~\ref{def:nc-test}) still has hope of holding, since it does not imply
learning a Bayes-optimal classifier.
Nevertheless, note that even the ``weak'' form is a fairly strong condition for neural networks: it implies that trained networks
(on \emph{any} size train set) learn feature-maps $h$ such that the push-forward $h_*(\cD)$ is a discrete measure.
Mapping the continuous measure $\cD$ to a discrete measure is a strong property, and one that is unlikely to hold for standard neural networks.

\paragraph{Feasibility of Weak-Collapse.}
While weak-collapse is unlikely to hold for neural networks trained with SGD, the definition itself is non-vacuous:
there exist
learning methods which are ``reasonable''
(asymptotically consistent) and exhibit weak test-collapse.
To see this, consider the following modified training procedure: first, train a neural network as usual to get a network $f : \mathcal{X} \to \mathcal{Y}$.
Then, construct another network $f'$ such that the \emph{last-layer representation} of $f'$ is
a one-hot encoding of the \emph{classification decision} of $f$.
That is, the representation $h'(x) \in \R^k$ satisfies $h'(x) := \vec{e}_{f(x)}$ where $\{\vec{e}_i\}$ are standard basis vectors.
This can be constructed by, for example, adding post-processing layers to $f$.
Now, the training procedure which outputs $f'$ will satisfy weak test-collapse of its representations, since its representations are always one of the $k$ standard basis vectors by construction.

\paragraph{Desirability of Neural Collapse for Generalization.}
Armed with these definitions, we can now consider whether
train or test collapse are necessary or sufficient
for on-distribution generalization.
First, neither train nor test collapse are strictly necessary for good generalization: As discussed, it is possible to construct models 
with identically good generalization performance, but which
satisfy neither train nor test collapse.
There are even natural, non-contrived examples of this:
models trained for less than one epoch
(the ``Ideal World'' in the terminology of \citet{nakkiran2020deep})
will not exhibit train collapse, because they are not trained to fit
their train set. And yet, as demonstrated in \citet{nakkiran2020deep},
they can match the performance of interpolating models.
This ``one epoch'' regime is also relevant in practice,
where models are trained on massive data sources such as internet scrapes, often for less than one epoch \citep{brown2020language,raffel2020exploring,komatsuzaki2019one}.

Further, neither train collapse (\cref{def:nc-train}) 
nor weak test-collapse (\cref{def:nc-test})
are sufficient for generalization.
It is possible to construct models which satisfy train collapse perfectly, but which are random functions at test time.
Likewise, it is possible to construct models which satisfy
weak test-collapse, but have random classification decisions.

Strong test-collapse (\cref{def:nc-test-strong})
\emph{is} sufficient for good test performance,
since it implies that test inputs map to the ``correct''
cluster in representation-space.
However, as we discussed,
strong test-collapse is infeasible, and impossible in practice.

\section{Experiments: Train and Test Collapse}
\label{sec:nc-experiments}
Here we complement our theoretical discussion by measuring
both train and test collapse in realistic settings, following
the experiments of \citet{Papyan24652}.
We find that train-collapse occurs in many settings, while
test-collapse (both strong and weak) does not.
We also show the dependency on the train set size:
larger train sets lead to \emph{stronger} test collapse, but \emph{weaker}
train collapse.
This further highlights the importance of distinguishing
between the two forms of collapse, since they can be anti-correlated in some settings.

\subsection{Measuring Collapse}
It is not possible to measure collapse
strictly according to
\cref{def:nc-train,def:nc-test,def:nc-test-strong},
since they involve a $t \to \infty$ limit.
Instead, we follow exactly the experimental procedure of
\citet{Papyan24652}, and measure approximations which
capture the ``degree of collapse.''
We restate their procedure here for convenience.
Measuring collapse require finding the vectors
$\mu_1, \mu_2, \dots \mu_k \in \R^d$, which embeddings collapse to.
The choice of these vectors depends on the setting, as below.

{\bf Train Collapse.}
For the train set, $\mu_i$ is defined as the train class-means:
\[
\hat{\mu}_i := 
\E_{(x, y) \in S}[ h^T_S(x) \mid y = i ]
\]
where $T$ is the maximum train time in the experiment.
Define the global mean as $\hat{\mu} := \sum_i \hat{\mu}_i / |\cY|$.
Then, the ``degree of train collapse''
is measured as:
\[
\textsf{TrainVariance}(t) :=
\frac{\E_{(x, y) \in S}[ ||h_S^t(x) - \hat{\mu}_y||^2 ]}
{ \E_i [|| \hat{\mu}_i - \hat{\mu}||^2 ] }
\]
Smaller values of this quantity indicate more ``collapse.''
The numerator here is the ``within-class variance''
and it is normalized by the ``between-class variance'',
in the terminology of \citet{Papyan24652}.
This definition follows the experimental measurements in \citet{Papyan24652}.

{\bf Strong Test Collapse.}
For test collapse, $\mu_i$ is defined as the test class-means:
\[
\bar{\mu}_i := 
\E_{(x, y) \sim \cD}[ h^T_S(x) \mid y = i ]
\]
The global mean is $\bar{\mu} := \sum_i \bar{\mu}_i / |\cY|$.
Then, the ``degree of strong test collapse''
is measured as:
\[
\textsf{StrongTestVariance}(t) :=
\frac{
\E_{(x, y) \sim \cD}[ ||h_S^t(x) - \bar{\mu}_y||^2 ]
}
{ \E_i [|| \bar{\mu}_i - \bar{\mu}||^2 ] }
\]

{\bf Weak Test Collapse.}
For weak test-collapse (\cref{def:nc-test}),
we do not require that representations collapse to their \emph{class means},
but simply to some $\mu_i$.
Thus, we define $\{\widetilde{\mu_i}\}$
as the result of $k$-means clustering on the following set of vectors:
\[
\{h^T_S(x)\}_{x \in \textrm{TestSet}}
\]
The global mean is $\widetilde{\mu} := \sum_i \widetilde{\mu}_i / |\cY|$.
And the ``degree of weak test collapse''
is measured as:
\[
\textsf{WeakTestVariance}(t) :=
\frac{\E_{(x, y) \sim \cD}[ \argmin_{i \in [k]} ||h_S^t(x) - \widetilde{\mu}_i||^2 ]}
{ \E_i [|| \widetilde{\mu}_i - \widetilde{\mu}||^2 ] }
\]

\subsection{Experimental Setup}
We briefly describe the setup for the train and test collapse experiments, including the datasets, modern architectures and training mechanisms.

\textbf{Datasets.} We consider image classification tasks with MNIST \citet{lecun1998gradient}, FashionMNIST \citet{xiao2017fashion}, CIFAR-10 \citet{krizhevsky2009learning}, SVHN \citet{netzer2011reading} and STL-10 datasets \citet{coates2011analysis}. SVHN was sub-sampled to $N=4600$ samples per class as training set and $N=1500$ samples per class for test set. Other datasets are following the standard setup. No data argumentation was done and we pre-process the images pixel-wise by subtracting the mean and dividing by the standard deviation. 

\textbf{Models.} We train standard Resnet18 and DenseNet201 for MNIST, FashionMNIST, CIFAR10 and SVHN. Resnet50 and DenseNet201 were trained for STL10. For all datasets we also train VGG11 with batch normalization. All models were trained from scratch with open source code from torchvision models. 

\begin{figure*}[t!]
\begin{center}
\includegraphics[width=\linewidth]{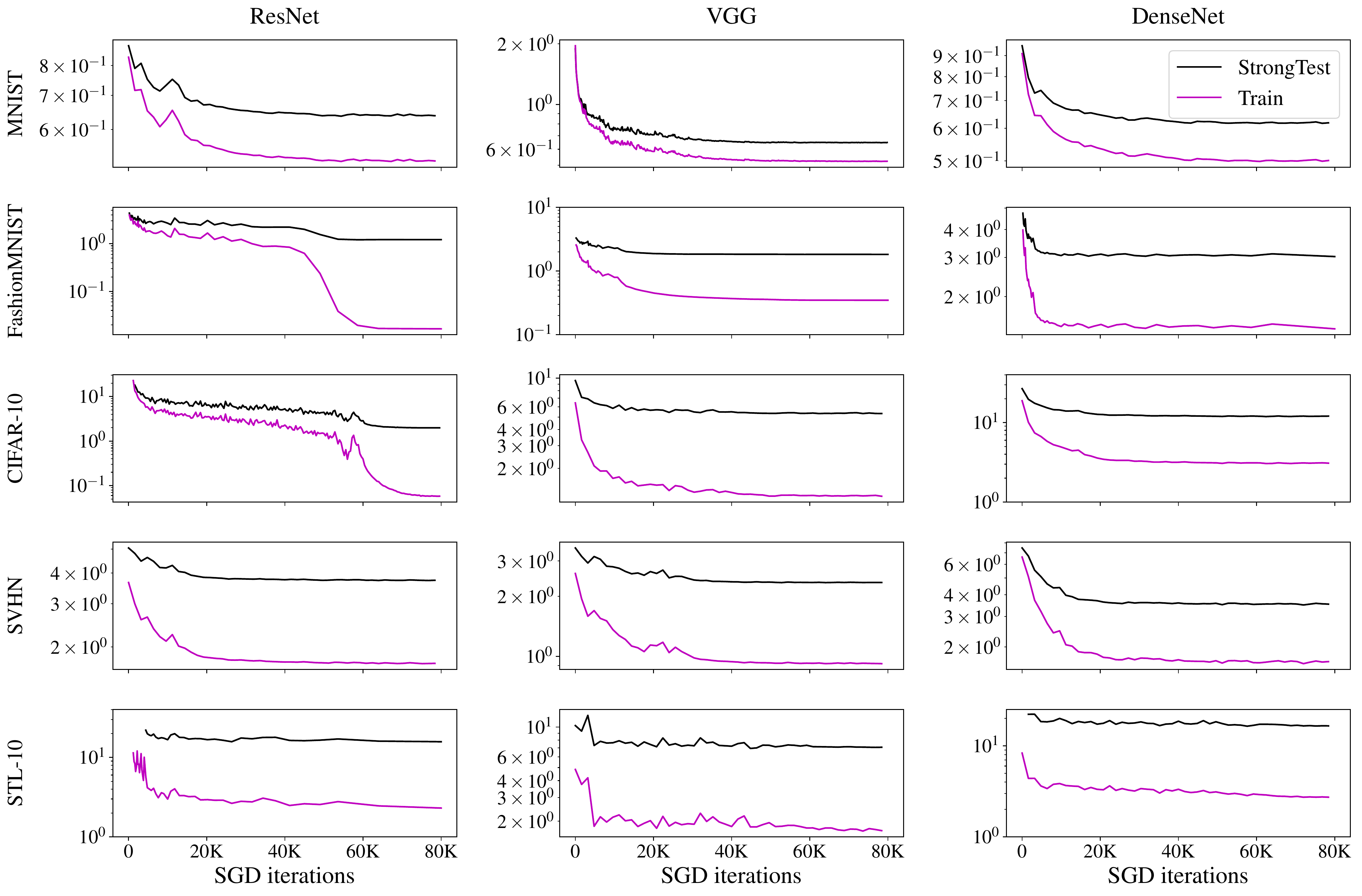}
\vskip -0.005in
\caption{{\bf Failure of Test Collapse.} Training and test variance vs. SGD iterations,
for various dataset and architecture combinations. We train all models to 0 training error and continue training to achieve close to 0 training loss.
All test sets (black line) do not collapse to
negligible variance,
and have much less collapse than the train sets (purple line).}
\label{fig:train_vs_test}
\end{center}
\vskip -0.2in
\end{figure*}
We use stochastic gradient descent (SGD) with momentum 0.9 and 
minimize the cross-entropy loss. All tasks were trained on a single GPU with batch size 128 and 80000 SGD iterations. Initial learning rate is 0.1 for Resnet18 and Resnet50 and 0.01 for DenseNet201 and VGG architectures. We decay the learning rate with cosine annealing scheme. 

\subsection{Results}
\begin{wrapfigure}{r}{0.4\textwidth}
\begin{center}
\includegraphics[width=\linewidth]{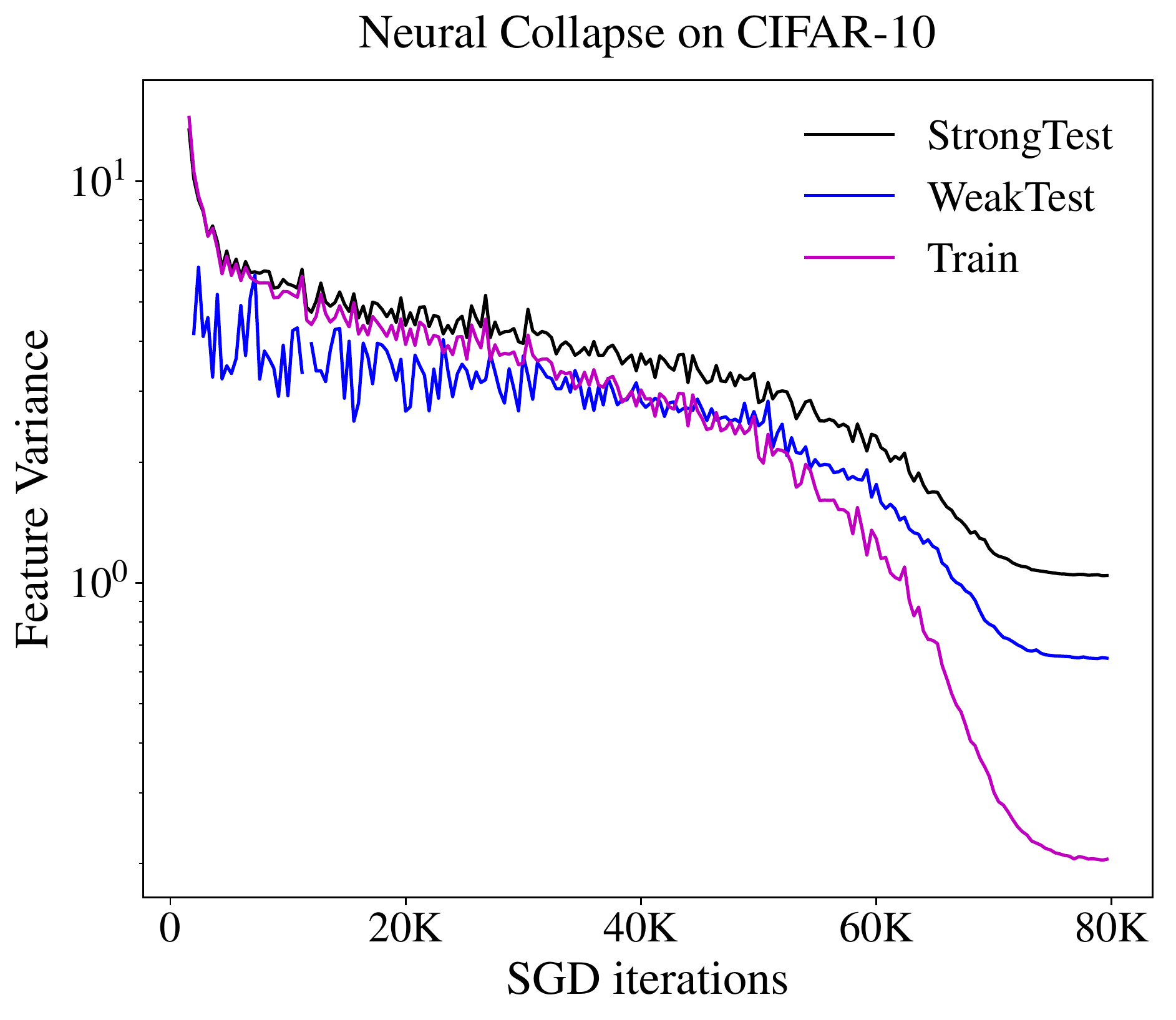}
\caption{{\bf Neural Collapse on CIFAR-10.}
Collapse occurs on the train set, but not on the test set
(neither Strong nor Weak). Weak test collapse has smaller variance than Strong test collapse.}
\label{fig:weak_collapse}
\end{center}
\vskip -0.6in
\end{wrapfigure}
In this section we show that the test collapse does not occur with experiments on a wide range of datasets and model architecture combinations.
We show that train collapse and test collapse
can be anti-correlated in some settings.

\paragraph{Failure of Test Collapse.}
In \cref{fig:weak_collapse}, we train a single model
(ResNet-18 on CIFAR-10)
and measure
\textsf{TrainVariance},
\textsf{WeakTestVariance},
and
\textsf{StrongTestVariance}
as a function of train time $t$.
That is, we measure the degree of train and test collapse
over increasing time.
We see that train collapse appears to occur,
while test collapse does not.
In particular, there is a ``generalization gap'' in the Train vs. Test Variances:
the
\textsf{TrainVariance}
appears to converge to 0 as $t \to \infty$,
while
\textsf{TestVariance} (both weak and strong) do not.
For the remainder of the experimental results,
we plot only ``strong'' test
collapse, since we generally observe that both strong and weak
collapse have similar behavior.

In \cref{fig:train_vs_test}, we train different models on various datasets and measure TrainVariance and StrongTestVariance as a function of train time $t$.
We train all models to get $0$ training error and continue training to achieve close to $0$ training loss\footnote{We 
use ``close to $0$'' to mean when the loss 
is below $10^{-5}$.}.
We see Strong Test Collapse does not occur on all settings, and has a large gap with Train Collapse. Again, the results show that Neural Collapse is mainly an optimization phenomenon and not a generalization one: test set does not collapse to negligible value in any setting.
As an aside, we observe that even TrainVariance is not always close to $0$, such as on CIFAR-10, SVHN and STL-10 datasets.
\begin{figure}[t!]
\begin{center}
\includegraphics[width=\linewidth]{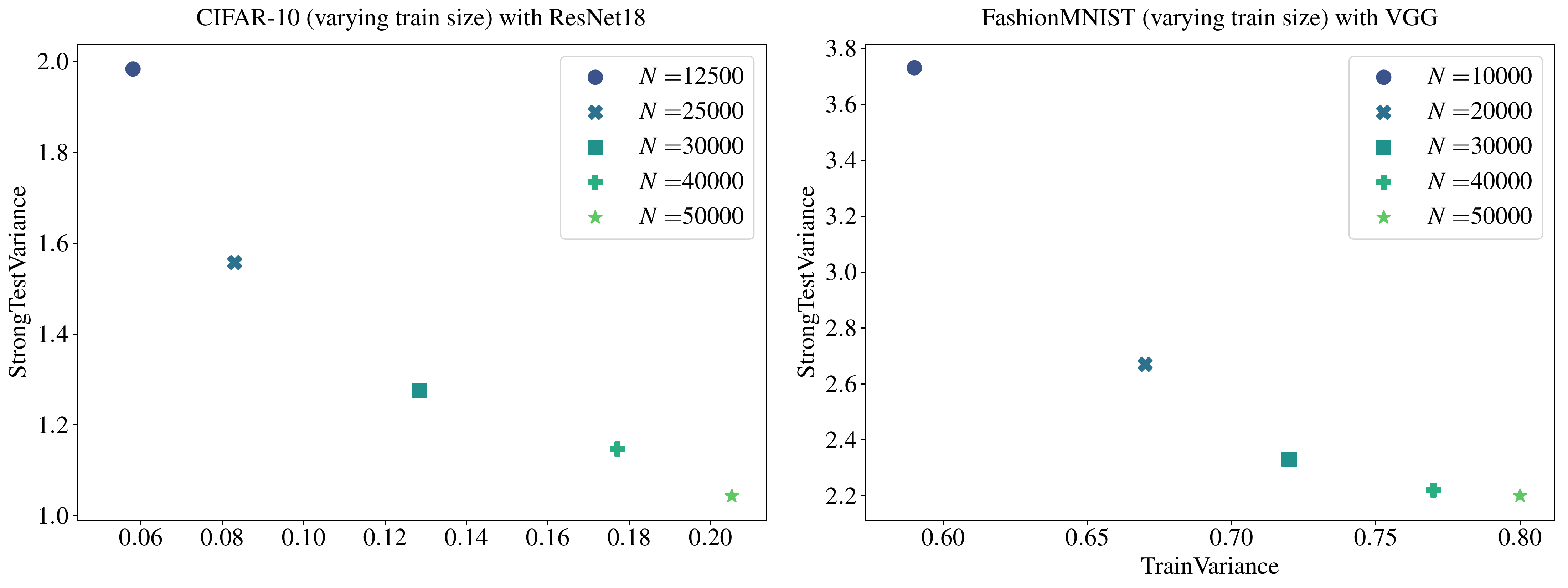}
\caption{{\bf Train vs. Test Anti-Correlation.} We vary the size of the train set ($N$), and observe that train and test collapse are anti-correlated.
\textit{Top:}  ResNet18 trained on subsets of CIFAR-10. \textit{Bottom:}  VGG11 trained on subsets of FashionMNIST.}
\label{fig:subset_collapse}
\end{center}
\vskip -0.20in
\end{figure}

\paragraph{Train vs. Test Anti-Correlation.}

In \cref{fig:subset_collapse} we train a ResNet18 on CIFAR-10, and vary the size of the train set from $N=12500, 25000, 30000, 40000$ to $N=50000$. We also report results on training a VGG11 network with batch normalization on different subsets of FashionMNIST.
We train all models past the point of $0$ training loss and note that in experiments of \cref{fig:subset_collapse} we stop training when the training loss decreases to $10^{-6}$.
\cref{fig:subset_collapse} plots the 
train collapse compared to the test collapse at the end of training,
for different train set sizes.
We find that as the train set size increases,
the test variation decreases (i.e. more test collapse),
while the train variation increases (less train collapse).
This illustrates that test and train collapse are not always correlated,
and thus it is important to distinguish between the two:
``better'' optimization behavior accompanies worse generalization behavior.

One limitation of this experiment is that we evaluate collapse at finite train time,
and not at $t = \infty$. 
Indeed, at $t = \infty$ we expect the train variation
to be identically $0$ for all data sizes
(by the definition of collapse),
but the test variation to decay with larger data sizes.
This situation is analogous to measuring train/test error itself for overparameterized models: for large enough models, train error will
always be $0$, but test error will decay with the data size.
This experiment thus highlights the 
importance of measuring both train \& test quantities,
and the subtlety involved in measuring collapse at finite time.

We also acknowledge that in this experiment,
increasing the size of the train set is correlated with
both better test collapse, and better generalization.
However, we caution that this should not be seen as
evidence that test collapse is mechanistically related
to generalization. First, because the test variance
does not truly ``collapse'', it just reduces, as already discussed.
And second, because this reduction in test variance is in some sense
necessary for any model with
improved test error--- since high test variance
would produce noisy classification decisions.
Thus, the correlation
of test variance and generalization in this
experiment should not be surprising.

\section{Collapsed Features Transfer Worse}
\label{sec:feature-learning}
In the previous section, we showed
that train-time collapse 
can be anti-correlated with generalization performance,
when measuring generalization on-distribution.
Now we investigate generalization on downstream tasks,
to understand the role of Neural Collapse in \emph{transfer-learning}
and \emph{representation learning}.

\subsection{Test Collapse implies Bad Representations}
We first observe that, using our definition of test collapse,
a model which has fully test-collapsed will have representations
that are bad for most downstream tasks.
To see this, consider the following example.
Suppose we have a distribution $\cD$ with 
ten types of images (as in CIFAR-10),
but we group them into two superclasses, such as ``animals'' and ``objects.''
We then train a classifier on this binary problem (e.g. CIFAR-10 images
with these binary labels).
Let the feature map of the fully-trained model
(that is, the limiting model as $t \to \infty$) be denoted $h$.
If this model exhibits even weak test collapse,
then there exist vectors $\{\mu_1, \mu_2\}$ such that the representations satisfy:
\begin{equation}
\label{eqn:repr}
\Pr_{x \sim \cD}\left[ h(x) \in \{\mu_1, \mu_2\} \right] = 1.
\end{equation}
That is, the representations will by definition ``collapse'':
every input $x \sim \cD$ will map to exactly one of two points $\mu_1, \mu_2$.
This property is clearly undesirable for representation learning.
For example, suppose we use these representations 
for learning on a related task: the original 10-way classification problem.
It is clear that no classifier using the fixed representations from $h$
can achieve more than 20\% test accuracy on the original 10-way task:
each group of 5 classes will collapse to a single point
after passing through $h$
(by \cref{eqn:repr}),
and will become impossible to disambiguate
among these 5 classes.
This shows that test collapse is undesirable for even an extremely
simple transfer learning task (where we transfer to the same distribution,
with finer label structure).
In the following sections, we will demonstrate almost this exact example
through experiments.

\begin{figure*}[!htbp]
  \centering
  \begin{minipage}[b]{.45\columnwidth}
    \includegraphics[width=\columnwidth]{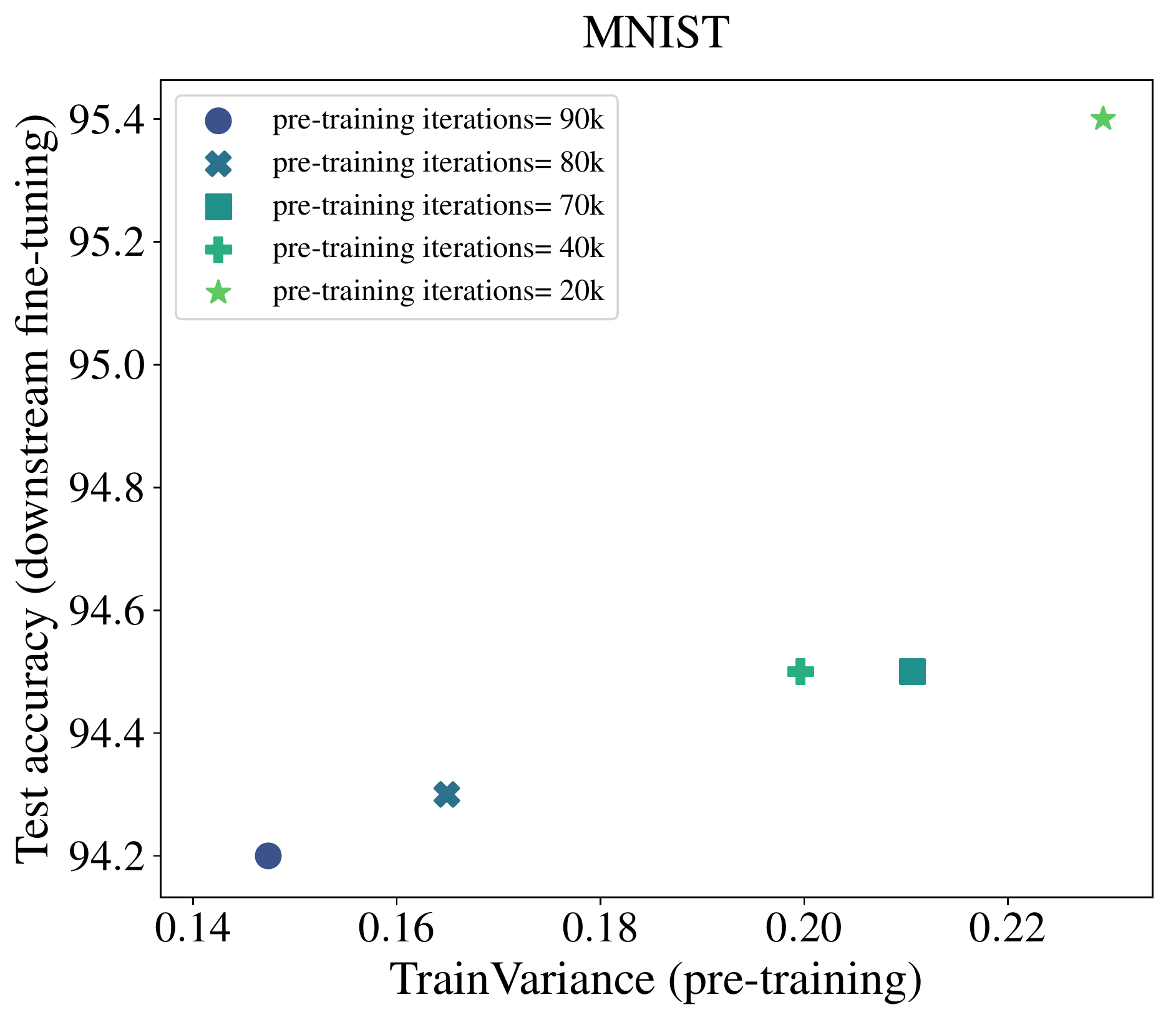}
  \end{minipage}
  \hfill
  \begin{minipage}[b]{.45\columnwidth}
    \includegraphics[width=\columnwidth]{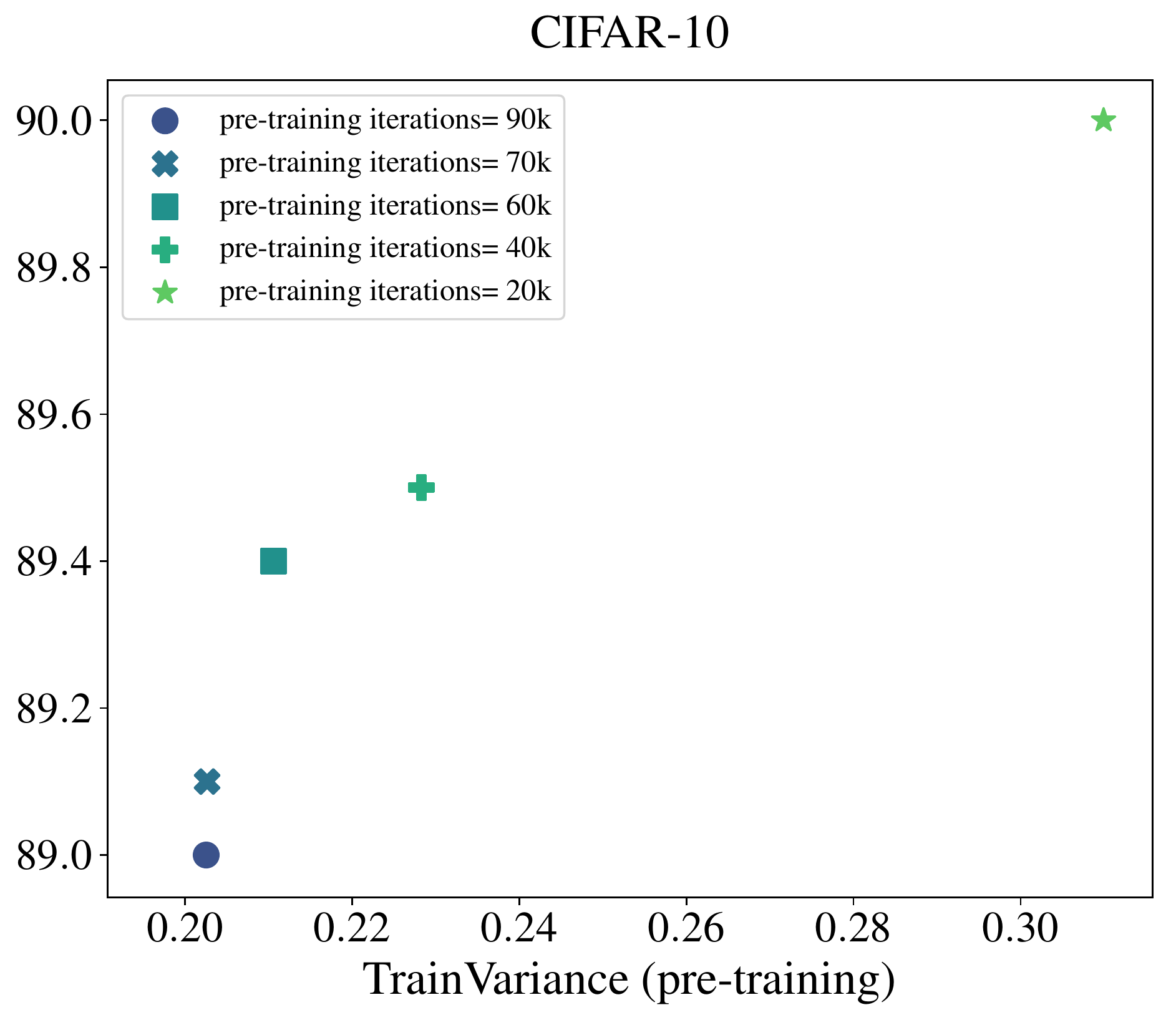}
  \end{minipage}
  \caption{{\bf Collapsed Features Transfer Worse.}
  We save different checkpoints during pre-training,
  and use them to initialize the downstream models. 
  The $x$-axis shows the TrainVariance
  of those checkpoints on the pre-training train set,
  and $y$-axis shows the test accuracy after fine-tuning on downstream tasks. 
  We find that stronger train collapse (i.e. lower variance) is correlated
  with lower downstream test accuracy.
  \textit{Left:} MNIST with a 3 hidden layer fully-connected network. \textit{Right:} CIFAR-10 with a standard Resnet18.}
  \label{fig:variance-accs}
\end{figure*}
\subsection{Experimental Setup}
 The transfer learning setup follows the standard pre-training and fine-tuning scheme. We train a 3 hidden layer fully-connected networks with 1024 units per layer on MNIST, and a standard Resnet18 on CIFAR-10.
 For pre-training, we use a subset of the train set and perform 2-class classification  (via super-classing). For \textit{fine-tuning}, we use the weights pre-trained as initialization of the weights other than the last classification layer, and do standard (10-class for MNIST, and 8-class for CIFAR-10) classification with a held-out subset.
 We do not report results with 
 linear probing,
 as it gives much worse transfer-performance than \textit{fine-tuning} scheme. See more details below.
 
\textbf{Super-class pre-training.} For MNIST, we set all odd numbers as one class and all even numbers as the other class. We train the model with the first $N=1000$ training samples as train set and the first $N=200$ test samples as test set. 

For CIFAR-10, we combine samples of `airplane, automobile, ship, truck' as one (objects) class and `bird, cat, frog, horse' as the other (animals) class. The two classes are balanced and have $40000$ training samples, and $8000$ test samples. We use a subset with $N=20000$ training samples (to keep each class balanced, we randomly choose $2500$ samples from each original class) and $N=4000$ ($500$ samples from each original class) test samples for pre-training.

The learning rate for MNIST with fully-connected networks is $0.001$ while for CIFAR-10 with ResNet18 is $0.1$. We decay learning rate with cosine annealing scheme. The models were trained minimizing the cross-entropy loss using SGD with momentum $0.9$ for $100000$ SGD iterations.

\textbf{Fine-tuning.} We initialize the weights (other than the last classification layer) of the downstream task with the pre-trained weights and fine-tune the whole network. For MNIST, we do the standard 10-class classification, while we sample another $500$ samples from training set for training and $100$ samples from the test set for inference. For CIFAR-10 we implement a 8-class classification (`airplane, automobile, ship, truck, bird, cat, frog, horse') with another $10000$ training samples as train set and another $2000$ test samples as test set. 

The optimization methodology is the same as in pre-training, other than the learning rate. We search over 0.0005 to 0.25 in fine-tuning for both MNIST and CIFAR-10 and report the best test accuracy of all swept learning rates.

\subsection{Results}
Here we show transfer learning results on MNIST and CIFAR-10.
As illustrated in \cref{fig:variance-accs}, we see that for both MNIST and CIFAR-10, the checkpoints with more Train Collapse gives worse
transfer-performance on downstream tasks.
That is, in these settings
more Train Collapse actually leads to learning worse features.
This demonstrates that neural collapse 
does not always lead to good representation learning--- and in some settings,
collapse actually harms representation quality.

\section{Cascading Collapse}
\label{sec:cascade}
Previous sections demonstrated that Neural Collapse is primarily an optimization phenomenon.
In this section, as an addendum, we present preliminary investigations which extend the Neural Collapse conjectures
by discussing optimization dynamics of even earlier layers (beyond the last layer).

\begin{wrapfigure}{r}{0.4\textwidth}
\centering
\vspace{-0.2in}
\includegraphics[width=0.85\linewidth]{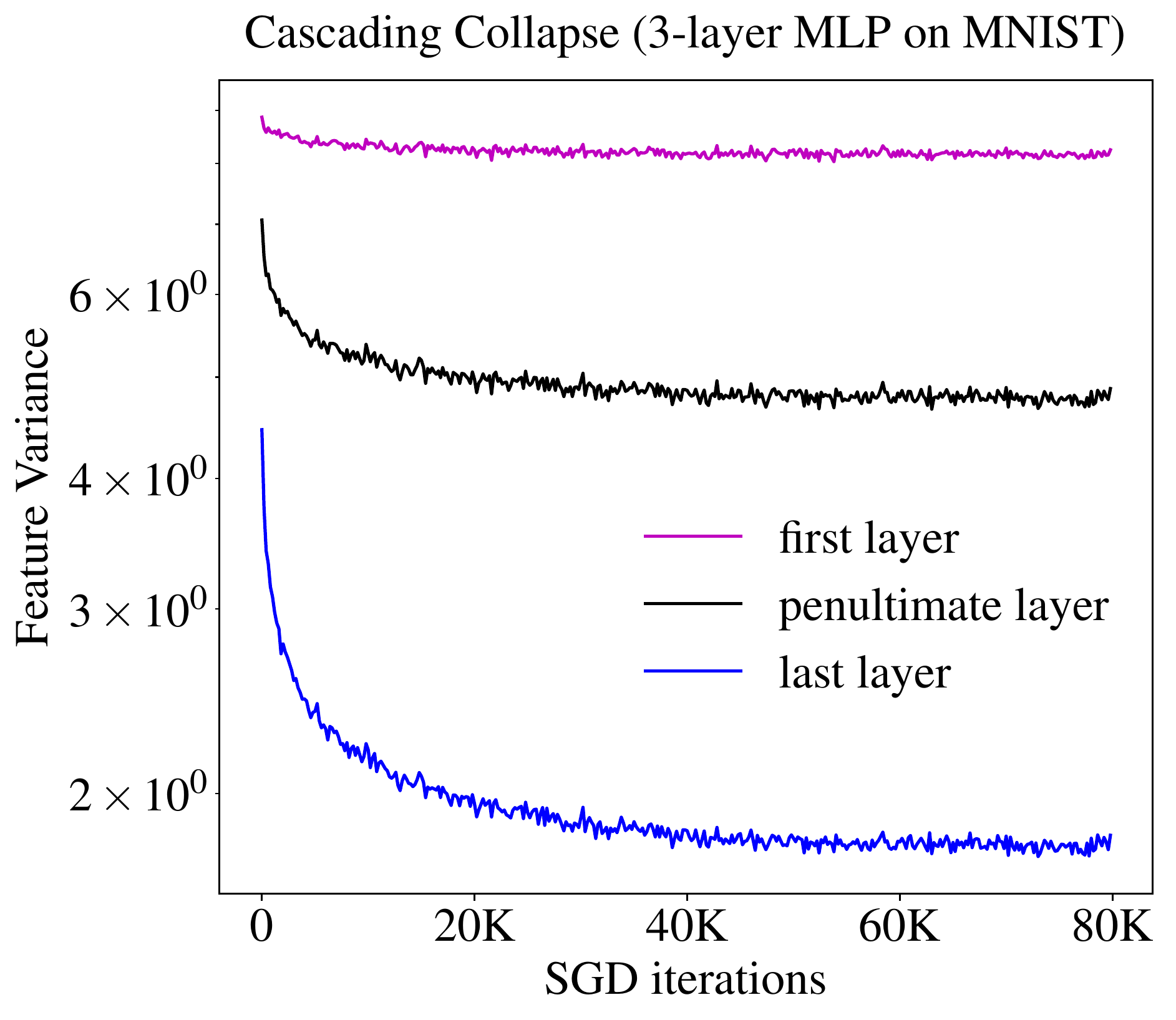}
\caption{{\bf Cascading collapse.}}
\label{fig:cascading}
\vspace{-0.2in}
\end{wrapfigure}
Existing work on Neural Collapse has focused on collapse of the last-layer features.
However, for sufficiently deep networks, the mechanisms driving
last-layer collapse could potentially extend to preceding layers as well.
Specifically, we know 
that at some late stage in training, the last layer exhibits Train Collapse
and remains essentially fixed for the remainder of training.
After this stage, we may expect that this fixed last-layer drives
a ``collapse'' in the preceeding layer features, and then in the layer before that,
and so on.
Here we give preliminary experimental evidence for such a ``cascading collapse,'' 
as an optimization phenomenon.
We do not formally define cascading collapse nor establish it conclusively--- this section
is meant only to suggest a potentially interesting avenue of future work.
 
\subsection{Experiments}
\textbf{Setup.} We train a fully-connected network with 3 hidden layer ($1024$ units per layer) on a subset of MNIST dataset. The subset contains $1000$ training samples ($100$ per class) and $200$ test samples ($20$ per class).

We minimize the cross-entropy loss using SGD with momentum 0.9 for $80000$ iterations. The initial learning rate is $0.001$ and we decay the learning rate with cosine annealing scheme.  

\textbf{Results. } We report one preliminary result in \cref{fig:cascading}. We measure TrainVariance as a function of train time $t$.
The ``last layer'' refers to the last hidden layer before the classification head.
As can be seen in \cref{fig:cascading}, the last layer features appear to collapse first,
and has the greatest degree of collapse.
The penultimate layer collapses qualitatively later, and to a lesser extent than the last layer.
And the first layer collapses the least, during the time scales observed.
We thus call this qualitative pattern of collapse among layers \emph{cascading collapse}.

We acknowledge that this is a preliminary experiment, and there are many
details to be determined. For example, it is unclear whether earlier layers will collapse to $0$ variance
in the $t \to \infty$ limit, or whether they will ``saturate'' at some finite variance, as they appear to do in the finite-time experiments.
Further, the relative speeds and sequential ordering of the collapse, across layers, is unclear from this experiment.
However, we believe that better understanding the implications of Neural Collapse for internal layers
is a fruitful research area, and we hope this section inspires future work on cascading collapse.

\section{Conclusion}
We point out that Neural Collapse is primarily an optimization phenomenon, not a generalization one, by investigating the train collapse and test collapse on various dataset and architecture combinations.
We propose more precise definitions--- ``strong'' and ``weak'' Neural Collapse for both the train set and the test set--- and discuss their theoretical feasibility.
We show that while train collapse reliably occurs in many settings,
test collapse does not.
We also show that train collapse can be anti-correlated with 
test performance in both on-distribution and transfer learning settings.
Our theoretical formulations and empirical observations suggest 
that while neural collapse continues to be an intriguing phenomenon and a promising optimization research program,
its relevance to generalization may be limited.

\subsubsection*{Acknowledgements}
LH thanks X. Y. Han for sharing experimental details. 
We thank Tatsunori Hashimoto for useful feedback on an early draft.

We are grateful for support of
the NSF and the
Simons Foundation for the Collaboration on the Theoretical
Foundations of Deep Learning\footnote{\url{https://deepfoundations.ai/}}
through awards DMS-2031883 and \#814639. We also acknowledge NSF support through  IIS-1815697 and the TILOS institute (NSF CCF-2112665).
We thank Nvidia for the donation of GPUs.
This work used the Extreme Science and Engineering Discovery Environment (XSEDE, \citet{xsede}),  which is supported by National Science Foundation grant number ACI-1548562 and allocation TG-CIS210104.

\subsubsection*{Author Contributions}
LH conceived the project, designed and performed all experiments,
and contributed to the writing.
MB advised the project, and contributed to the editing.
PN managed the project, and contributed to the framing, writing, and theoretical formulation.

\newpage

\bibliography{refs}
\bibliographystyle{icml2022}

\end{document}